\documentclass[letterpaper]{article} 
\usepackage{aaai25}  

\usepackage{subcaption}
\usepackage{caption}
\usepackage{times}  
\usepackage{helvet}  
\usepackage{courier}  
\usepackage[hyphens]{url}  
\usepackage{graphicx} 
\usepackage{natbib}  
\usepackage{caption} 
\usepackage{booktabs}
\usepackage{newfloat}
\usepackage{listings}
\usepackage{amsmath}
\usepackage{algorithmic}
 \usepackage{multirow}
\usepackage{bm}      
\usepackage{amssymb}
\DeclareMathOperator*{\argmax}{arg\,max}

\newtheorem{definition}{Definition}

\newtheorem{proposition}{Proposition}
\newtheorem{corollary}{Corollary}

\DeclareCaptionStyle{ruled}{labelfont=normalfont,labelsep=colon,strut=off} 
\usepackage{algorithm}
\usepackage{algorithmic}
\usepackage{subcaption}
\usepackage{comment}
\usepackage{xcolor}
\usepackage{newfloat}
\usepackage{listings}
\DeclareCaptionStyle{ruled}{labelfont=normalfont,labelsep=colon,strut=off} 
\floatstyle{ruled}
\newfloat{listing}{tb}{lst}{}
\floatname{listing}{Listing}
\title{MAPLE: A Framework for Active Preference Learning \\ Guided by Large Language Models}
\author{
   Saaduddin Mahmud, ~Mason Nakamura, ~Shlomo Zilberstein
}
\affiliations{

University of Massachusetts Amherst\\ Amherst, Massachusetts, USA\\
\{smahmud, mnakamura, shlomo\}@umass.edu
}

\usepackage{bibentry}

\begin{document}

\maketitle

\begin{abstract}

The advent of large language models (LLMs) has sparked significant interest in using natural language for preference learning. However, existing methods often suffer from high computational burdens, taxing human supervision, and lack of interpretability. To address these issues, we introduce \textbf{MAPLE}, a framework for large language model-guided Bayesian active preference learning. MAPLE leverages LLMs to model the distribution over preference functions, conditioning it on both natural language feedback and conventional preference learning feedback, such as pairwise trajectory rankings. MAPLE also employs active learning to systematically reduce uncertainty in this distribution and incorporates a language-conditioned active query selection mechanism to identify informative and easy-to-answer queries, thus reducing human burden. We evaluate MAPLE's sample efficiency and preference inference quality across two benchmarks, including a real-world vehicle route planning benchmark using OpenStreetMap data. Our results demonstrate that MAPLE accelerates the learning process and effectively improves humans' ability to answer queries.

\end{abstract}

\section{Introduction}
Following significant advancements in artificial intelligence, autonomous agents are increasingly being deployed in real-world applications to tackle complex tasks~\citep{zilberstein2015building, dietterich2017steps}. A prominent method for efficiently aligning these agents with human preferences is Active Learning from Demonstration (Active LfD)~\citep{biyik2022learning}. Preference-based Active LfD, a variant of LfD, aims to infer a preference function from human-generated rankings over a set of observed behaviors using a Bayesian active learning approach.

Recent advancements in natural language processing have inspired many researchers to leverage language-based abstraction for learning human preferences~\citep{soni2022towards, guan2022leveraging}. This approach offers a more flexible and interpretable way to learn preferences compared to conventional methods~\citep{sadigh2017active,brown2019drex, browngoo2019trex}. More recent work~\citep{yu2023language,ma2023eureka} has focused on utilizing large language models (LLMs), such as ChatGPT~\citep{achiam2023gpt}, with prompting-based approaches to learn preferences from natural language instructions. However, these methods often require significant computational resources and taxing human supervision, as they lack a systematic querying approach. 

To tackle these challenges, we introduce a novel framework—MAPLE (\textbf{M}odel-guided \textbf{A}ctive \textbf{P}reference \textbf{Le}arning). MAPLE begins by interpreting natural language instructions from humans and utilizes large language models (LLMs) to estimate a distribution over preference functions. It then applies an active learning approach to systematically reduce uncertainty about the correct preference function. This is achieved through standard Bayesian posterior updates, conditioned on both conventional preference learning feedback, such as pairwise trajectory rankings, and linguistic feedback such as clarification or explanations of the cause behind the preference. To further ease human effort, MAPLE incorporates a language-conditioned active query selection mechanism that leverages feedback on the difficulty of previous queries to choose future queries that are both informative and easy to answer. MAPLE represents preference functions as a linear combination of abstract language concepts, providing a modular structure that enables the framework to acquire new concepts over time and enhance sample efficiency for future instructions. Moreover, this interpretable structure allows for human auditing of the learning process, facilitating human-guided validation before applying the preference function to optimize behavior.

In our experiments, we evaluate the efficacy of MAPLE in terms of sample efficiency during learning, as well as the quality of the final preference function. We use an environment based on the popular Minigrid~\cite{MinigridMiniworld23} and introduce a new realistic vehicle routing benchmark based on OpenStreetMap ~\cite{OpenStreetMap} data, which includes text descriptions of the road network of different cities in the USA. Our evaluation shows the effectiveness of MAPLE in preference inference and improving human's ability to answer queries. Our contributions are threefold:
\begin{itemize}
    \item We propose a Bayesian preference learning framework that leverages LLMs and natural language explanations to reduce uncertainty over preference functions.
    \item We provide a language-conditioned active query selection approach to reduce human burden.
    \item We conduct extensive evaluations, including the design of a realistic new benchmark that can be used for future research in this area.
\end{itemize}

\section{Related Work}

\paragraph{Learning from demonstration}
Most Learning from Demonstration (LfD) algorithms learn a reward function using expert trajectories~\citep{ng2000algorithms, abbeel2004apprenticeship, ziebart2008maximum}. Some of these approaches utilize a Bayesian framework to learn the reward or preference function~\citep{ramachandran2007bayesian, brown2020safe, rev}, and some pair it with active learning to reduce the number of human queries~\citep{sadigh2017active, Basu2018LearningFR, biyik2022learning}. However, these methods are unable to utilize natural language abstraction, whereas our method can use both. In addition, we employ language-conditioned active learning to reduce user burden, an approach not previously explored in this context.

\paragraph{Natural language in intention communication}
With the advent of natural language processing, several works have focused on directly communicating abstract concepts to agents~\cite{tevet2022human, guo2022generating, wang2024rl, sontakke2024roboclip, lin2022inferring, tien2024optimizing, lou2024safe}. The key difference is that these works directly condition behavior on natural language, whereas we learn a language-abstracted preference function. This approach offers several advantages, including increased transparency, a more fine-grained trade-off between concepts, and enhanced transferability. The work most closely related to ours is \cite{lin2022inferring}, which infers rewards from language but restricts them to step-wise decision-making.

Other lines of work~\cite{yu2023language, ma2023eureka} aim to learn reward functions directly by prompting LLMs. However, these methods are limited by the variables available in the coding space and often struggle with identifying temporally extended abstract behaviors. Further, these approaches can not utilize conventional preference feedback, whereas MAPLE can utilize both. Additionally, they either lack a systematic way of acquiring human feedback or rely on data-hungry evolutionary algorithms. In contrast, our approach employs more efficient Bayesian active learning.

\paragraph{Abstraction in reward learning}
Several works leverage abstract concepts to learn reward functions~\citep{lyu2019sdrl, illanes2020symbolic, icarte2022reward, guan2022relative, soni2022towards, bobu2021learning, guan2021widening, guan2022leveraging, silver2022learning, zhangdual, bucker2023latte, cui2023no}. Two methods closely related to our work are PRESCA~\citep{soni2022towards} and RBA~\citep{guan2022relative}. PRESCA learns state-based abstract concepts to be avoided, while RBA learns temporally extended concepts with two variants: global (eliciting preference weights directly from humans) and local (tuning weights using binary search). Our approach also leverages temporally extended concepts but learns preference functions from natural language feedback using active learning. Unlike RBA, which relies on direct preference weights from humans or binary search, our method uses LLM-guided active learning
for more expressive and informative preference elicitation, thereby reducing human effort. 

Some works use offline behavior datasets or demonstrations to learn diverse skills~\citep{lee2010learning, wang2017robust, zhou2018cost, peng2018sfv, luo2020carl, chebotar2021actionable, 2021-TOG-AMP}, which complement our approach. While MAPLE can also utilize such datasets in pre-training, the focus of MAPLE is to encode human preference in terms of these concepts using natural language.

\paragraph{Alignment auditing}
Alignment auditing ensures that an agent's behavior aligns with human intentions by verifying that the agent has learned the correct preference function. While some works focus on alignment verification with minimal queries~\citep{Brown2021ValueAV}, they often rely on function weights, value weights, or trajectory rankings, which are difficult to interpret. In contrast, our approach leverages natural language to communicate with humans, facilitating validation and serving as a stopping criterion for the active learning process. \citet{rev} presents a notable alignment auditing approach related to our method, using explanations to detect misalignment and update distributions over preferences. While they employ a feature attribution method, we use natural language explanations. Additionally, they use human-selected or randomly sampled data points from an offline dataset for auditing, whereas we employ active learning to enhance efficiency.

\paragraph{Active learning}
Previous works have explored different acquisition functions for active learning, typically focusing on selecting queries that maximize certain uncertainty quantization metrics. These metrics include predictive entropy~\citep{gal2016dropout}, uncertainty volume reduction~\citep{sadigh2017active}, mutual information maximization~\citep{biyik2019asking}, and maximizing variation ratios~\citep{gal2016dropout}. Our approach complements these methods by integrating language-conditioned query selection to reduce user burden. While any of these methods can be paired with MAPLE, we opt for variation ratio due to its ease of calculation and high effectiveness.

\begin{figure*}[t]
    \centering
    \includegraphics[width=0.90\textwidth]{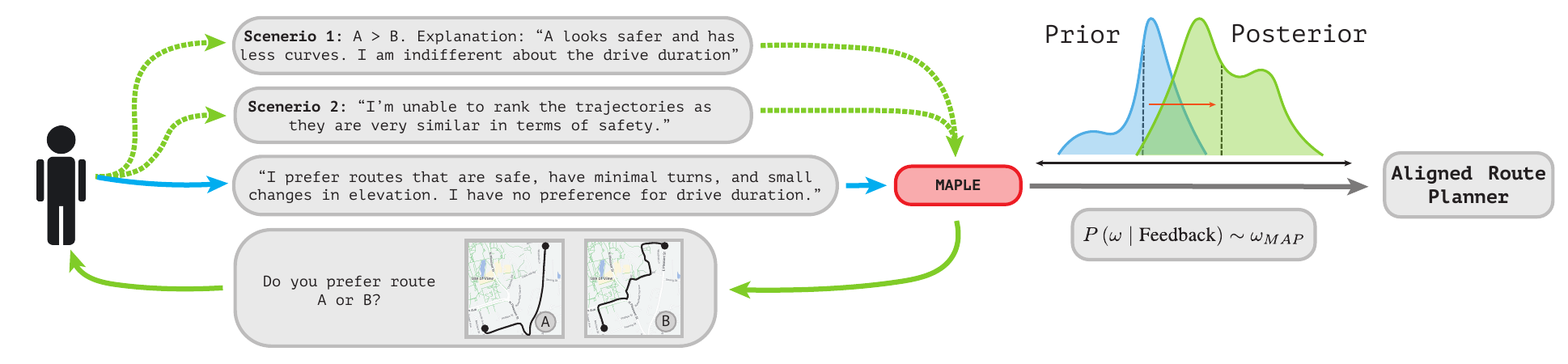}
    \caption{Application of MAPLE to the Natural Language Vehicle Routing Task.}
    \label{fig:motivation}
\end{figure*}

\section{Background}

\paragraph{Markov decision process (MDP)} A Markov Decision Process (MDP) \( M \) is represented by the tuple \( M = (S, A, T, S_0, R, \gamma) \), where \( S \) is the set of states, \( A \) is the set of actions, \( T: S \times A \times S \rightarrow [0,1] \) is the transition function, \( S_0 \) is the initial state distribution, and \( \gamma \in [0,1) \) is the discount factor. A history \( h_t \) is a sequence of states up to time \( t \), $(s_0, \dots, s_t)$. The reward function \( R: H \times A \rightarrow [-R_{\text{max}}, R_{\text{max}}] \) maps histories and actions to rewards. For some problems, a goal function \( G: H \rightarrow [0,1] \) is provided that maps histories to goal achievements. In such problems, the reward function is typically \( R: H \times A \rightarrow [-R_{\text{max}}, 0] \) and \( \forall a \in A \), \( T(s_g, a, s_g) = 1\) and \( R(h_t \cup s_g, a) = 0 \) given the final state \( s_g \in h_t \). A policy \( \pi: H \times A \rightarrow [0,1] \) is a mapping from histories to a distribution over actions. The policy \( \pi \) induces a value function \( V^\pi: S \rightarrow \mathbb{R} \), which represents the expected cumulative return \( V^\pi(s) \) that the agent can achieve from state \( s \) when following policy \( \pi \). An optimal policy \( \pi^* \) maximizes the expected cumulative return \( V^*(s) \) from any state \( s \), particularly from the initial state \( s_0 \).

\paragraph{Bayesian preference learning} A preference function \( \omega \) maps a trajectory \( \tau \) to a real number reflecting the alignment of the trajectory with the human's objective. The goal of preference learning is to infer this function from various types of human feedback. A common approach involves learning this function from a pairwise preference dataset, denoted by \( \mathcal{D} = \{ (\tau^1_1 \succ \tau^2_1), (\tau^1_2 \succ \tau^2_2), \ldots, (\tau^1_n \succ \tau^2_n) \} \), where \(\tau^1_i\) and \(\tau^2_i\) are two different trajectories, and \(\tau^1_i \succ \tau^2_i\) indicates that \(\tau^1_i\) is preferred to \(\tau^2_i\). A Bayesian framework for preference learning, as described in \citet{ramachandran2007bayesian}, defines a probability distribution over preference functions given a trajectory dataset \(\mathcal{D}\) using Bayes' rule: \( P(\omega \mid \mathcal{D}) \propto P(\mathcal{D} \mid \omega)P(\omega)\). Various algorithms define \( P(\mathcal{D} \mid \omega) \) differently, but we adopt the definition from BREX \citep{brown2020safe} using the Bradley–Terry model~\cite{Bradley1952RankAO}:\\[-8pt]
\begin{equation}
P(\mathcal{D} \mid \omega) = \prod_{(\tau^1_i \succ \tau^2_i) \in \mathcal{D}} \frac{e^{\beta \omega(\tau^1_i)}}{e^{\beta \omega(\tau^1_i)} + e^{\beta \omega(\tau^2_i)}}.
\label{BT}
\end{equation}
Here, \( \beta \in [0, \infty) \) is the inverse-temperature parameter.

\paragraph{Variance ratio}
Given a conditional probability distribution $P(\cdot \mid X)$ over $\{y_i\}_{i=0}^k$, the variance ratio of an input $X$ is defined as follows:\\[-3 pt]
$$
    \text{Variance\_Ratio}(X) = 1-\argmax_{y_i} P(y_i \mid X)
$$  
\section{Problem Formulation}

\paragraph{MAPLE}
We define a MAPLE problem instance as the tuple \((M_{-R}, C, \Omega, D_\tau, \mathcal{H}, \mathbb{L})\), where:
\begin{itemize}
    \item \(M_{-R}\) is an MDP with an undefined reward function \( R \).
    \item \(\mathcal{H}\) is the human interaction function that acts as the interface between the human and the MAPLE framework. Humans provide their feedback, preferences, and explanations in response to natural language queries posed by MAPLE.
    \item \(\mathbb{L}\) is the LLM interaction function that generates natural language queries to the LLM and returns structured output in text files, such as JSON format.
    \item \(C\) is an expanding set of natural language concepts \(\{ c_1, c_2, \ldots, c_n\} \). We also use \( C(\cdot) \) to refer to a mapping model that takes a trajectory embedding \( \phi(\tau) \) and a natural language concept embedding \( \psi(c_i) \) and maps them to a numeric value indicating the degree to which the trajectory \( \tau \) satisfies the concept \( c_i \). For non-Markovian concepts, \( \phi(\cdot) \) may be a sequence model such as a transformer. For Markovian concepts, we can define \( C(\phi(\tau), \psi(c_i)) = \sum_{\displaystyle s \in \tau} C(\phi(s), \psi(c_i)) \), where \( \phi(s) \) is the state embedding.
    \item \(\Omega\) is the space of all preference functions. In MAPLE, the preference functions \( \omega \) over a trajectory $\tau$ are modeled as a linear combination of the concepts and their associated weights:\\[-3pt]
    \begin{equation}
        \omega(\tau) = \sum_{\displaystyle c_i \in C} \omega_{c_i} \cdot C(\phi(\tau), \psi(c_i))
    \end{equation}
    \item \(D_\tau\) is a dataset of unlabeled trajectories \(\{\tau_1, \tau_2, \ldots, \tau_m\}\).
\end{itemize}

The objective of MAPLE is to model the repeated interaction between a human and an agent, where the human communicates their task objective \(\mathcal{A}^{\mathcal{H}}_{\mathcal{T}}\) in natural language, and the agent is responsible for completing the task in alignment with that objective. MAPLE accomplishes this by actively learning a symbolic preference function \(\omega\) using large language models (LLMs), enabling the agent to optimize its behavior according to this function to ensure its actions align with human preferences.

\paragraph{Motivating example}
Consider an intelligent route planning system that takes a source, a destination, and user preferences about the route in natural language, as illustrated in Figure\,\ref{fig:motivation}. Datasets for several preference-defining concepts such as speed, safety, battery friendliness, smoothness, autopilot friendliness, and scenic view can be easily obtained and used to pre-train the concept mapping function \( C(\cdot) \). The goal of MAPLE is to take natural language instructions from a human and map them to a preference function \( \omega \) interactively so that a search algorithm can optimize it to find the preferred route. MAPLE incorporates preference feedback on top of natural language feedback to address issues like hallucination and calibration associated with directly using LLMs. Additionally, MAPLE allows the human to skip difficult queries and learns in-context which query to present, making the system more human-friendly. Furthermore, the preference function inference process in MAPLE is fully interpretable, enabling a human to audit the process thoroughly and provide the necessary feedback for improvement. Finally, the interaction with the human is repeated, allowing MAPLE to acquire new concepts over time and become more efficient for future tasks.

\section{Detailed Description of the Proposed Method}
A key innovation of MAPLE is the integration of conventional feedback from the preference learning literature with more expressive linguistic feedback, formally captured within a Bayesian framework introduced in REVEALE~\cite{rev}:
\begin{equation}
    P(\omega \mid F_h, F_l) \propto P(F_h \mid \omega)P(F_l \mid \omega)P(\omega)
\end{equation}
Above, \( F_h \) represents the set of feedback observed in conventional preference learning algorithms, specifically in the context of this paper pairwise trajectory ranking.\footnote{MAPLE can handle any conventional feedback for which $P(F_h \mid \omega)$ is defined.} \( F_l \) denotes the set of linguistic feedback. We can rewrite the equation as:
\begin{align}
P(\omega \mid F_h, F_l) &\propto \underbrace{P(F_h \mid \omega)}_{\text{Bradley-Terry Model}} \underbrace{P(\omega \mid F_l)}_{\text{LLM}} \underbrace{P(F_l)}_{\text{Uniform}}\\[-4pt]
&\propto \underbrace{P(F_h \mid \omega)}_{\text{Bradley-Terry Model}} \underbrace{P(\omega \mid F_l)}_{\text{LLM}}
\end{align}

Here, the likelihood of \( F_h \) given \( \omega \) is defined using the Bradley-Terry Model. The likelihood of \( \omega \) given \( F_l \) is estimated using an LLM. Beyond incorporating linguistic feedback via LLMs, MAPLE advances conventional active learning methods. Conventional active learning typically focuses on selecting queries that reduce the maximum uncertainty of the posterior but lacks a flexible mechanism to account for human capability in responding to certain types of queries. MAPLE's Oracle-guided active query selection enhances any conventional acquisition function by leveraging linguistic feedback to alleviate the human burden associated with difficult queries. In the rest of this section, we provide more details on MAPLE, particularly Algorithms~\ref{alg:maple} and \ref{alg:lqs}.

\begin{algorithm}[t]
\scriptsize
\caption{MAPLE}
\begin{algorithmic}[1]  
\REQUIRE Human instruction $\mathcal{A}^{\mathcal{H}}_{\mathcal{T}}$, 
Acquisition function $\mathcal{A}_f$, \# of LLM query $K$
\STATE $F_h, F_q \leftarrow \emptyset, \emptyset$
\STATE $F_l \leftarrow \{\mathcal{T}\}$
\STATE $\Omega_{\mathcal{T}} \leftarrow \{\omega_i\}_{i=0}^n \sim \mathbb{L}(\omega \mid F_l) $
\WHILE{condition not met}
    \STATE $Q \leftarrow \{(\tau_i,\tau_j): \tau_i,\tau_j \in \mathcal{D}_{\tau} \land (\tau_i,\tau_j) \not\in F_h \}$
    \STATE $q \leftarrow$ Query Selection($\mathcal{A}_f$, $Q$, $F_q$, $\Omega_{\mathcal{T}}$, $\mathbb{L}$, $K$)
    \STATE $(f_h,f_l,f_q) \leftarrow \mathcal{H}(q)$
    \STATE $F_h, F_l, F_q \leftarrow F_h \cup \{f_h\}, F_l \cup \{f_l\}, F_q \cup \{f_q\}$
    \STATE $\Omega_{\mathcal{T}} \leftarrow \{\omega_i\}_{i=0}^n \sim P(F_h \mid \omega)P(\omega \mid F_l)$
\ENDWHILE

\STATE \textbf{return} $\Omega_{\mathcal{T}}$

\end{algorithmic}
\label{alg:maple}

\end{algorithm}

\subsection{Initialization}
MAPLE starts by taking natural language instruction about task preference \(\mathcal{A}^{\mathcal{H}}_{\mathcal{T}}\) and initializes the pairwise preference feedback set $F_h$, linguistic feedback $F_l$, and feedback about query difficulty $F_q$ (lines 1-2, Algorithm~\ref{alg:maple}). After that, the initial set of weights is sampled using the LLM from the distribution $P(\omega \mid F_l)$ as $F_h$ is still empty (line 3, Algorithm~\ref{alg:maple}). 
To sample $\omega$ from $P(\omega \mid F_l)$ we explore two sampling strategies described below.
\paragraph{Preference weight sampling from LLM}
We directly prompt the LLM \(\mathbb{L}\) to provide linear weights \(\omega\) over the abstract concepts. Specifically, we provide \(\mathbb{L}\) with a prompt containing the task description \(\mathcal{T}\), a list of known concepts \(C\), human preference \(\mathcal{A}^{\mathcal{H}}_{\mathcal{T}}\), and examples of instruction weight pairs, along with additional answer generation instructions \(G\) (see Appendix for details). The LLM processes this prompt and returns an answer \(\mathcal{A}^{\mathbb{L}}_{\omega_i}\):\\[-6pt]
\[
\mathcal{A}^{\mathbb{L}}_{\omega_i} \leftarrow \mathbb{L}(\text{prompt}(\mathcal{T}, C, \mathcal{A}^{\mathcal{H}}_{\mathcal{T}}, D_{\mathcal{I}}, G)) 
\]

We can take advantage of text generation temperature to collect a diverse set of samples. We define the set of all generated weights as \(\mathcal{A}^{\mathbb{L}}_{\omega} \). Then $P(\omega_j \mid F_l)$ can be modeled for any arbitrary $\omega_j$ as follows:
\begin{equation}
     P(\omega_j \mid F_l) = \exp\left(-\beta_{l} \mathbb{E}_{\omega_i \in \mathcal{A}^{\mathbb{L}}_{\omega}} [\text{Distance}(\omega_i,\omega_j)]\right)
\label{eq:lw1}
\end{equation}
In this case, Euclidean or Cosine distance can be applied.
\paragraph{Distribution weight sampling using LLM}
The second approach we explore is distribution modeling using an LLM. Here, we use similar prompts as in the previous approach; however, we instruct the LLM to generate parameters for $P(\omega \mid F_l)$. For example, for the weight of each concept $\omega_{c_i} \in \omega$ we prompt \(\mathbb{L}\) to generate a range \(\omega_{c_i}^{\text{range}} = [\omega_{c_i}^{\text{min}}, \omega_{c_i}^{\text{max}}]\). Then we can define $P(\omega \mid F_l)$ as follows:
\begin{small}
\setlength{\abovedisplayskip}{3pt} \setlength{\abovedisplayshortskip}{3pt}
\begin{equation}
    P(\omega \mid F_l) =
    \begin{cases}
        1, & \text{if } \omega_{c_i} \in \omega_{c_i}^{\text{range}}, \forall \omega_{c_i} \in \omega  \\
        
        0, & \text{otherwise}.
    \end{cases}
    \label{eq:lw2}
\end{equation}
\end{small}
We can similarly model this for other forms of distributions, such as the Gaussian distribution. Once the initialization process is complete, MAPLE iteratively reduces its uncertainty using human feedback.

\subsection{LLM-Guided Active Preference Learning}
After initialization, MAPLE iteratively follows three steps: 1) query selection, 2) human feedback collection, and 3) preference
posterior update, discussed below.

\paragraph{Oracle-guided active query selection (OAQS)} At the beginning of each iteration, MAPLE selects a query $q$ (a pair of trajectories) (lines 5-6, Algorithm\,\ref{alg:maple}) from $\mathcal{D}_{\tau}$ that would reduce uncertainty the most while mitigating query difficulty based on human feedback. The query selection process is described in Algorithm\,\ref{alg:lqs}, which starts by sorting all the queries based on an acquisition function $\mathcal{A}_f$. In this paper, we use the variance ratio for its flexibility and high efficacy. In particular, for trajectory ranking queries, the score for \((\tau_i,\tau_j)\) is calculated as $\mathbb{E}_{\omega \sim \Omega_{\mathcal{T}}}[1 - \max(P(\tau_i \succ \tau_j \mid \omega), P(\tau_j \succ \tau_i \mid \omega))]$. Note that other acquisition functions can also be used. Once sorted, OAQS iterates over the top $K$ queries and selects the first query that the oracle (in our case an LLM) evaluates to be answerable by the human (lines 2-11). Finally, Algorithm\,\ref{alg:lqs} returns the least difficult query $q$ among the top K query selected by $\mathcal{A}_f$. We now analyze the performance of OAQS based on the characterization of the oracle.\footnote{Proofs are in the Appendix.} 

\begin{definition}
Let \( Q \) denote the set of all possible queries, and \( Q_{\mathcal{A}} \subseteq Q \) represent the subset of queries answerable by \( \mathcal{H} \). The \textbf{Absolute Query Success Rate (AQSR)} is defined as the probability that a randomly selected query \( q \) belongs to the intersection \( Q \cap Q_{\mathcal{A}} \), i.e., \( P(q \in Q_{\mathcal{A}}) \).
\end{definition}

\begin{definition}
The \textbf{Query Success Rate (QSR)} of a query selection strategy is defined as the probability that a query \( q \), selected by the strategy, belongs to \( Q_{\mathcal{A}} \), i.e., \( P(q \in Q_{\mathcal{A}} \mid \text{strategy}) \).
\end{definition}

\begin{proposition}
Assuming the independence of AQSR from acquisition function ranking, the QSR of a \textbf{random query selection strategy}: \( P(q \in Q_{\mathcal{A}} \mid \text{random}) = AQSR \)
\end{proposition}

\begin{proposition}
Under the same assumption of proposition 1, the QSR of a \textbf{top-query selection strategy}, which always selects the highest-rated query by $\mathcal{A}_f$, \( P(q \in Q_{\mathcal{A}} \mid \text{top}) = AQSR \).
\end{proposition}

\begin{proposition}
The QSR of the \textbf{OAQS} strategy is given by \\[-4pt]
\[
AQSR \cdot Y_1 \cdot \frac{1 - \left[ \text{AQSR} \cdot (1 - Y_0 - Y_1) + Y_0 \right]^K}{1 - \left[ \text{AQSR} \cdot (1 - Y_0 - Y_1) + Y_0 \right]},
\]
where \( Y_0 = P(\mathbb{L}(F_q, q \notin Q_{\mathcal{A}}) = \text{False}) \) and \\\( Y_1 = P(\mathbb{L}(F_q, q \in Q_{\mathcal{A}}) = \text{True}) \). Here, we assume independence of AQSR, \( Y_0 \), and \( Y_1 \) from acquisition function ranking.
\label{prop1}
\end{proposition}

\begin{corollary}
Based on Proposition \ref{prop1}, the OAQS will have a higher QSR than the random query selection strategy and top-query selection strategy iff, \( Y_0 + Y_1 > 1 \) as \( K \rightarrow \infty \).
\end{corollary}

\begin{definition}
The \textbf{Optimal Query Success Rate (OQSR)} of a strategy is defined as the probability that the strategy returns the query \( q^* \) with the highest value according to an acquisition function \( \mathcal{A}_f \), among all answerable queries, i.e., \\[-7pt]
\[
P(q^* = \arg\max_{q \in Q} \mathcal{A}_f(q) \mathbb{I}(q \in Q_{\mathcal{A}})),
\]
where \( q^* \) is the query returned by the strategy.
\end{definition}

\begin{proposition}
Under the similar assumption of proposition 1. assumption, the OQSR of a \textbf{random query selection strategy} is equal to $1/|Q|$.
\end{proposition}

\begin{proposition}
Under the similar assumption of proposition 1, the OQSR of a \textbf{Top-Query Selection Strategy} is equal to the AQSR.
\end{proposition}

\begin{proposition}
Under the same assumption of Proposition 3, the OQSR of the \textbf{OAQS} strategy is given by\\[-8pt]
\[
\text{OQSR} = AQSR \cdot Y_1 \cdot \frac{1 - [(1 - AQSR)Y_0]^K}{1 - (1 - AQSR)Y_0}.
\]
\label{prop6}
\end{proposition}

\begin{corollary}
Based on Proposition \ref{prop6}, the OAQS strategy will have a higher OQSR than the top-query selection strategy if \( (1-AQSR)Y_0 + Y_1 > 1 \) as \( K \rightarrow \infty \), and then random query selection strategy if \( \mathrm{AQSR} \cdot Y_1 > \frac{1 - (1 - \mathrm{AQSR}) Y_0}{|Q|} \) as \( K \rightarrow \infty \).
\end{corollary}

\paragraph{Human feedback collection} MAPLE queries the human \(\mathcal{H}\) using the query $q$ returned by Algorithm\,\ref{alg:lqs} to collect feedback. For each query $q$, MAPLE provides a pair of trajectories, and \(\mathcal{H}\) returns an answer $\mathcal{A}^{\mathcal{H}}_{\tau} = (f_h, f_l, f_q)$, where $f_h$ is binary feedback, $f_l$ is an optional natural language explanation associated with that feedback---possibly empty if the human does not provide an explanation---and $f_q$ is a optional natural language feedback about the difficulty of the query. Each piece of feedback is then added to the corresponding feedback set (lines 7-8, Algorithm\,\ref{alg:maple}). 

\paragraph{LLM-guided posterior update} Once feedback is added to the set, we update our current weight sample $\Omega_{\mathcal{T}}$ by sampling $P(F_h \mid \omega)P(\omega \mid F_l)$ using MCMC sampling, where $P(\omega \mid F_h)$ is given by Equation~\ref{BT}, and $P(\omega \mid F_l)$ is given by Equations~\ref{eq:lw1} and \ref{eq:lw2}.

\paragraph{Stopping criteria} MAPLE can employ various stopping criteria for active query generation, including:
\begin{itemize}
    \item A fixed budget approach, where MAPLE operates within a predefined maximum query limit.
    \item A human-gated stopping criterion, based on the human's assessment of the system's competence. MAPLE's interpretability enhances this process, allowing the inclusion of its current predictions and explanations in each query for human evaluation (line 7, Algorithm\,\ref{alg:maple}).
\end{itemize}

\begin{algorithm}[t]
\scriptsize
\caption{Oracle-Guided Query Selection}
\begin{algorithmic}[1]  
\REQUIRE Acquisition function $\mathcal{A}_f$, List of queries $Q$, Query preference feedback $F_q$, Set of weights from current posterior $\Omega_{\mathcal{T}}$, Oracle $\mathcal{O}$, \# of Oracle queries $K$
\STATE $Q_{sort} \leftarrow \text{sort}(Q|\mathcal{A}_f, \Omega_{\mathcal{T}})$ 
\STATE $Q_{top} \leftarrow Q_{sort}[0:K] $
\FOR{$q \in Q_{top}$}
    \STATE $s_q \leftarrow \mathcal{O}(\text{prompt}(F_q, q))$
    \IF{$s_q$ is True}
        \STATE \textbf{return} $q$ 
    \ENDIF
\ENDFOR
\STATE \textbf{return} $Q_{sort}[0]$ 
\end{algorithmic}
\label{alg:lqs}
\end{algorithm}

\paragraph{Handling unknown concepts} It should be noted that humans may provide instructions $\mathcal{A}^{\mathcal{H}}_{\mathcal{T}}$ that cannot be sufficiently captured by the available concepts in the concept maps. While this case is beyond the scope of this paper, several remedies exist in the literature to address this issue. LLMs can be prompted to add new concepts when generating weights. By leveraging the generalization capability of $C(\cdot)$ we can attempt to apply these new concepts directly. If the new concept is significantly different from those in $C$, few-shot-learning techniques can be employed. In particular, during interactions, if a new concept is important, we can use non-parametric few-shot learning from human feedback, such as nearest neighbor search, to improve concept mapping~\cite{tian2024survey}. Finally, if a new concept arises repeatedly, it can be added to the concept map by retraining $C$ with data collected from multiple interactions through few-shot learning, as considered in~\cite{soni2022towards}.

\subsection{Policy Optimization}
The method for utilizing the weights generated by MAPLE to optimize policy varies based on the trajectory encoding and the chosen policy solver algorithm. For example, for Markovian preferences, the weights can be directly used with an MDP solver. In non-Markovian settings, the weights can be used to rank trajectories and directly align the policy with algorithms such as DPO~\cite{rafailov2024direct}, or train a dense reward function~\cite{guan2022relative} using preference learning algorithms such as TREX~\cite{browngoo2019trex}, and then use that reward function with reinforcement learning algorithms.

\begin{figure}[t]
    \centering
        \centering
        \includegraphics[width=3in]{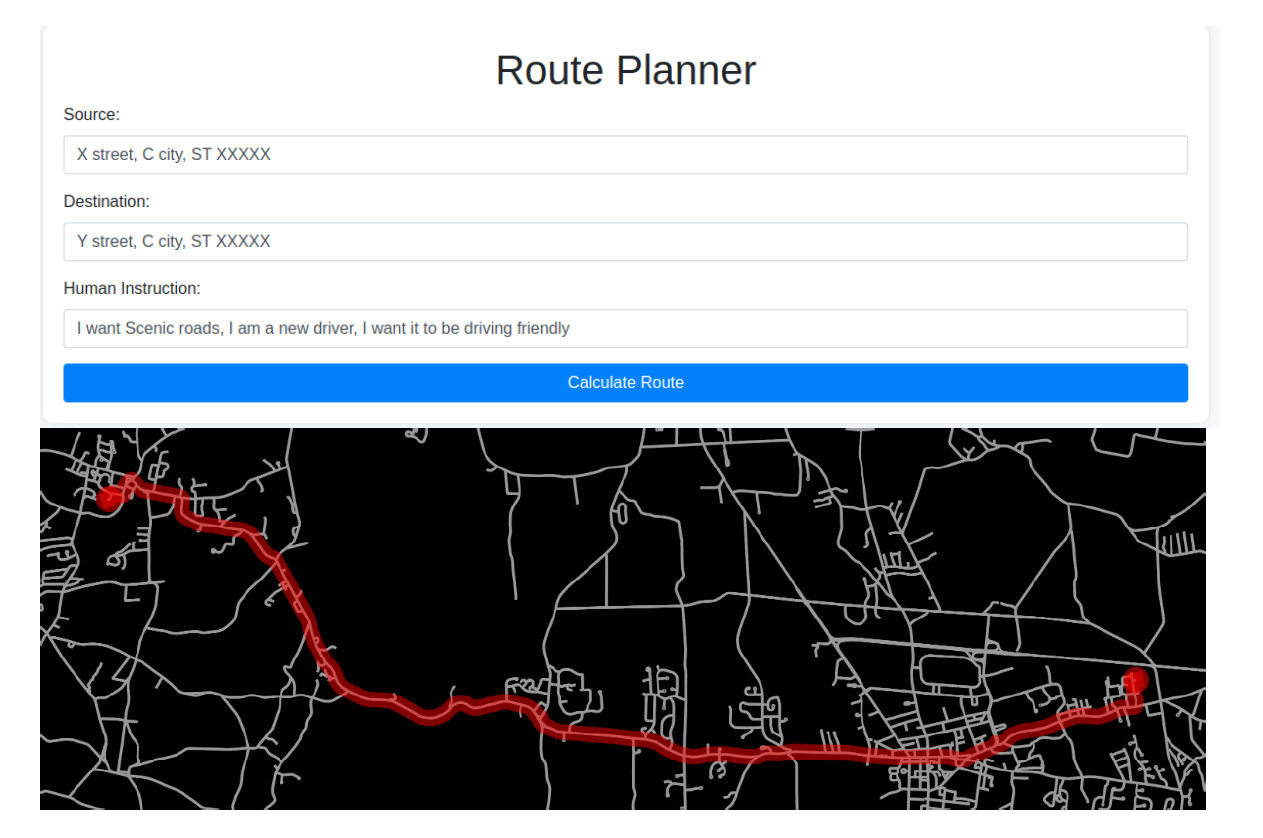}
         \caption{OpenStreetMap Routing}
        \label{fig:image1}
\end{figure}

\section{Experiments}

In this section, we describe a comprehensive evaluation of MAPLE within the two environments detailed below. It is important to note that none of the models used in our experiments were fine-tuned; they were utilized in their publicly available form. We ran the local language model, specifically Mistral-7B-instruct-v0.3 (4-bit quantization), on a computer equipped with 64GB RAM and an Nvidia RTX 4090 24GB graphics card. For larger models, we relied on public API infrastructure. Note that we present results using preference weight sampling as it outperformed distribution weight sampling in both benchmarks (Appendix for details). 

\paragraph{OpenStreetMap Routing}
We use OpenStreetMap to generate routing graphs for different U.S. states. The environment includes a concept mapping function capable of using ten different concepts: 1) Time, 2) Speed, 3) Safety, 4) Scenic, 5) Battery Friendly, 6) Gas Station Nearby, 7) Charging Station Nearby, 8) Human Driving Friendly, 9) Battery ReGen Friendly, and 10) Autopilot Friendly. The goal is to find a route between a given source and destination that aligns with user preferences. To generate $D_{\tau}$, we used 200 random source and destination pairs with randomly sampled weights from $\Omega$. For modeling human interaction, we utilized two different datasets, each containing 50 human interaction templates. The first dataset, called ``Clear,'' provides clear, knowledgeable instructions. The second dataset, called ``Natural,'' obfuscates the ``Clear'' dataset with more natural-sounding language typical of everyday conversation and contextual information, for example:
\begin{quote}
\small
    \textbf{Clear:} ``I prefer routes that are safe and scenic, with a moderate focus on speed and low importance on time.''
\end{quote}
\begin{quote}
\small
    \textbf{Natural:} ``I'm planning a weekend drive to enjoy the countryside, so I'm not in a hurry. I want the route to be as safe as possible because I'll be driving with my family. It would be great if the drive is scenic too, so we could take in the beautiful views along the way. Speed isn't a top concern, and we're really just out to enjoy the journey rather than worry about how long it takes to get there.''
\end{quote}
For modeling $f_l$, the human clarifies the type of car (gas, autonomous, or electric) with a probability of 0.2 per feedback. For $f_q$, the human is unable to answer when the top two highest-weighted (based on ground-truth weights) concepts in both trajectories are closer than a predefined threshold.
\begin{figure}[!t]
    
    \centering
    \includegraphics[height=5cm]{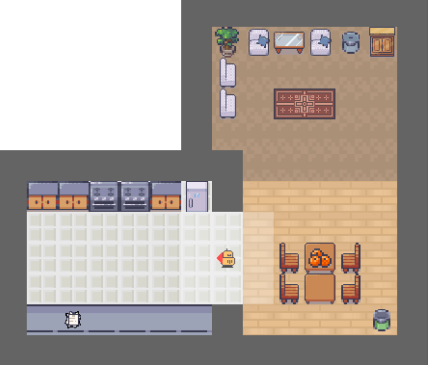}
    \caption{HomeGrid}
    \label{fig:image2}

\end{figure}

\paragraph{HomeGrid}
The HomGrid environment is a simplified Minigrid~\cite{MinigridMiniworld23} setting designed to simulate a robot performing household tasks~\cite{lin2023learning}. It features a discrete, finite action space and a partially observable language observation space for a $3\times 3$ grid, detailing the objects and flooring in each grid square, within a truncated $12\times 14$ grid. The initial abstract concepts include: 1) avoiding objects such as tables and chairs, 2) avoiding walls, 3) avoiding placing objects like bottles and plates on the floor, 4) avoiding placing objects on the stove, and 5) avoiding placing objects on the left chairs. A total of 60 trajectories were manually generated to update the posterior distribution of the weights $\omega$ for each method. For modeling $f_l$, the human highlights the concept that was most influential for their preference.  The modeling of $f_q$ follows a similar approach to that used in OSM Routing.
\subsection{Experimental Results}
\begin{figure*}[t]
    \centering
    \vspace{-10pt}  
    \begin{minipage}[b]{0.30\textwidth}
        \centering
        \includegraphics[width=0.8\textwidth]{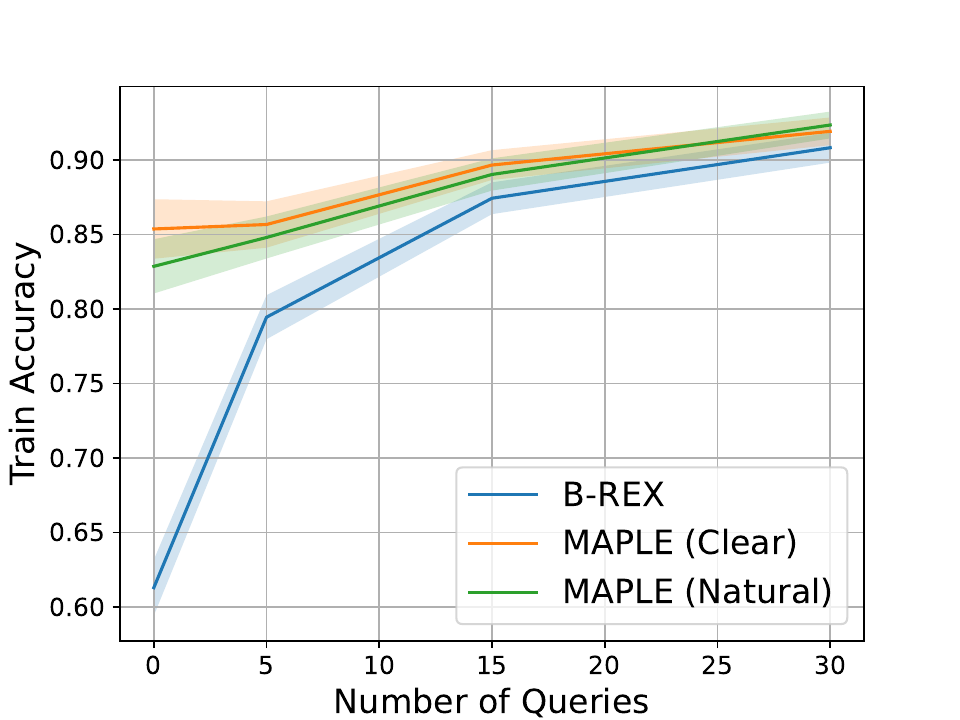}
        \caption*{(a) Test accuracy (OSM Routing)}
    \end{minipage}
    \begin{minipage}[b]{0.30\textwidth}
        \centering
        \includegraphics[width=0.8\textwidth]{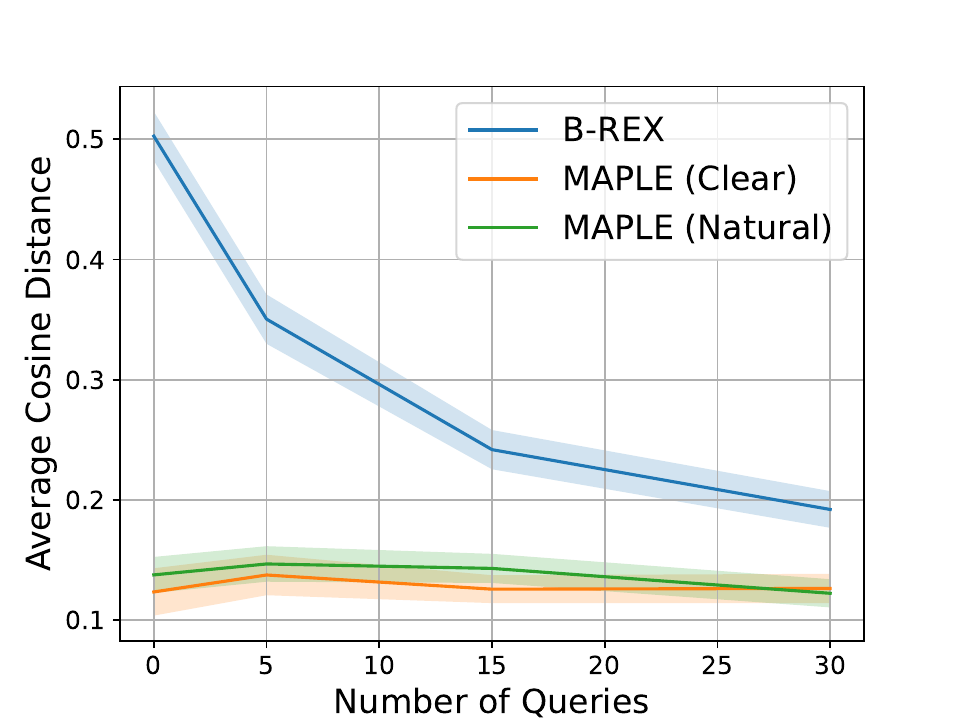}
        \caption*{(b) Cosine distance (OSM Routing)}
    \end{minipage}
    \hfill
    \begin{minipage}[b]{0.36\textwidth}
        
        \centering
        \tiny
        
        \begin{tabular}{@{}lcc@{}}
        
            \toprule
            & \multicolumn{2}{c}{\textbf{OSM Routing}} \\ 
            \textbf{Model} & \textbf{Test Accuracy} & \textbf{Expected Cost $\Delta$} \\ \midrule
            B-REX           & $0.79 \pm 0.04$ & $0.66 \pm 0.10$ \\
            Mistral-7B-Ins. & $0.85 \pm 0.02$ & $0.34 \pm 0.07$ \\
            GPT-4o          & $\mathbf{0.88 \pm 0.01}$ & $\mathbf{0.22 \pm 0.05}$ \\
            GPT-4o-mini     & $0.87 \pm 0.02$ & $0.29 \pm 0.07$ \\
            Gemini-1.5-Pro  & $0.86 \pm 0.01$ & $0.26 \pm 0.05$ \\
            Mistral-Large-2 & $0.85 \pm 0.02$ & $0.27 \pm 0.06$ \\ \bottomrule
        \end{tabular}
        \caption*{(c) Test accuracy and cost for different models with 5-feedback (OSM Routing, Natural)}
    \end{minipage}

    \begin{minipage}[b]{0.30\textwidth}
        \centering
        \includegraphics[width=0.8\textwidth]{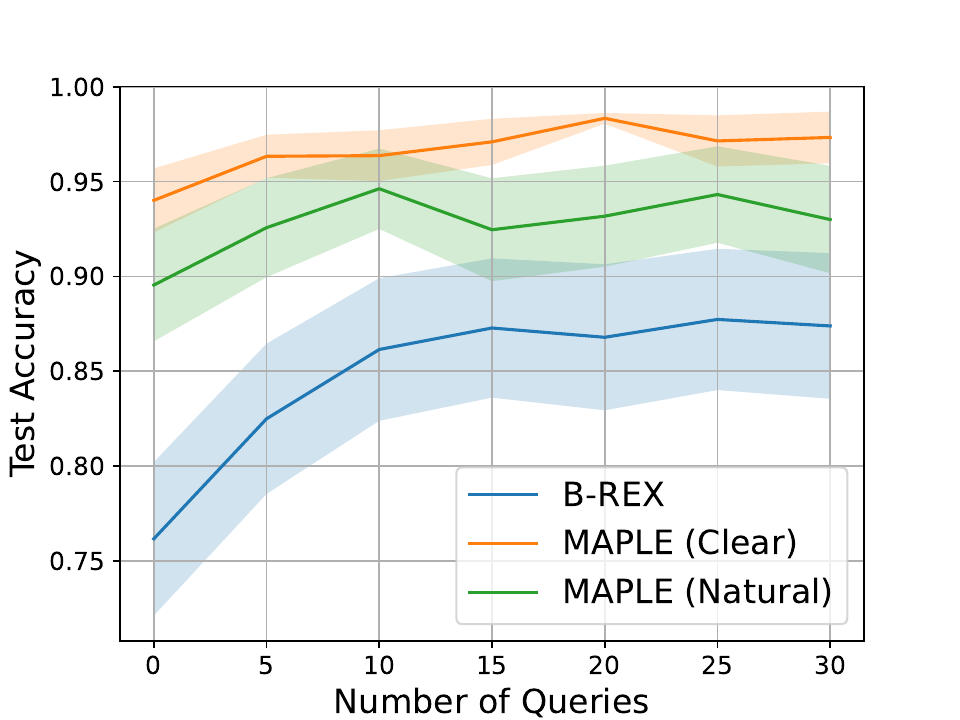}
        \caption*{(d) Test accuracy (HomeGrid)}
    \end{minipage}
    \begin{minipage}[b]{0.30\textwidth}
        \centering
        \includegraphics[width=0.8\textwidth]{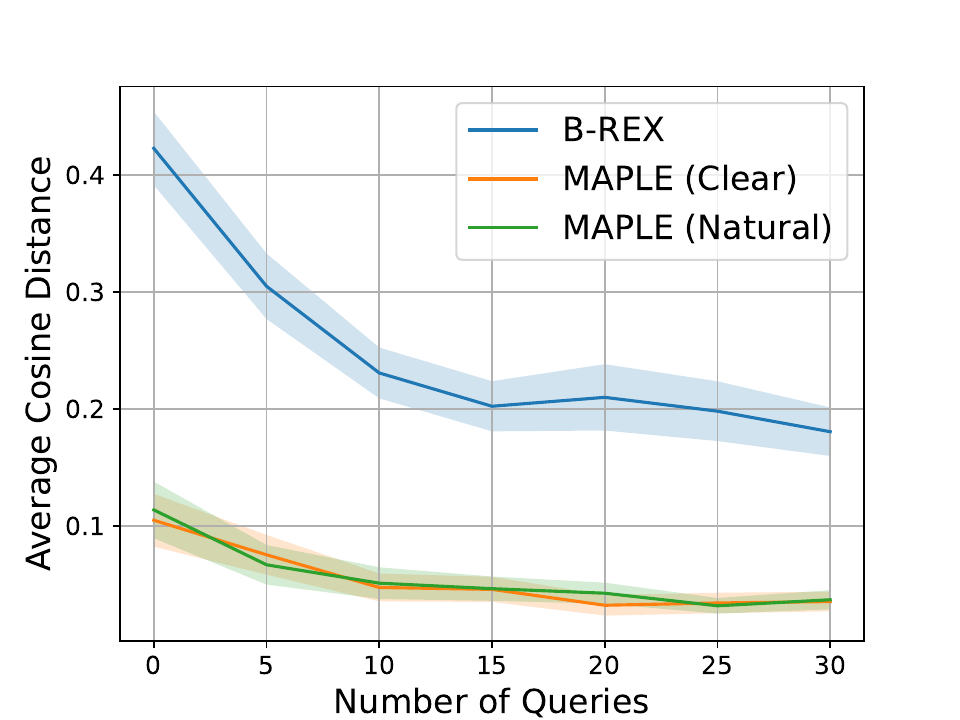}
        \caption*{(e) Cosine distance (HomeGrid)}
    \end{minipage}
    \hfill
    \begin{minipage}[b]{0.36\textwidth}
        \centering
        \tiny
        \begin{tabular}{@{}lcc@{}}
            \toprule
            & \multicolumn{2}{c}{\textbf{HomeGrid}} \\ 
            \textbf{Model} & \textbf{Test Accuracy} & \textbf{Expected Cost $\Delta$} \\ \midrule
            B-REX           & $0.82 \pm 0.04$ & $1.40 \pm 0.24$ \\
            Mistral-7B-Ins. & $\mathbf{0.92} \pm \mathbf{0.02}$ & $0.67 \pm 0.13$ \\
            GPT-4o          & $0.87 \pm 0.03$ & $0.89 \pm 0.20$ \\
            GPT-4o-mini     & $0.83 \pm 0.04$ & $3.90 \pm 0.66$ \\
            Gemini-1.5-Pro  & $0.87 \pm 0.04$ & $1.00 \pm 0.18$ \\
            Mistral-Large-2 & $0.85 \pm 0.04$ & $\mathbf{0.55 \pm 0.14}$ \\ \bottomrule
        \end{tabular}
        \caption*{(f) Test accuracy and cost for different models with 5-feedback (HomeGrid, Natural)}
    \end{minipage}
    \vspace{-5pt}
    \caption{Comparison of efficacy of language feedback for preference inference.}
    \label{fig:combined}
\end{figure*}

\begin{figure*}[!t]
    \centering
    \vspace{-8pt}  
    \begin{minipage}[b]{0.3\textwidth}
        \centering
        \includegraphics[width=0.8\textwidth]{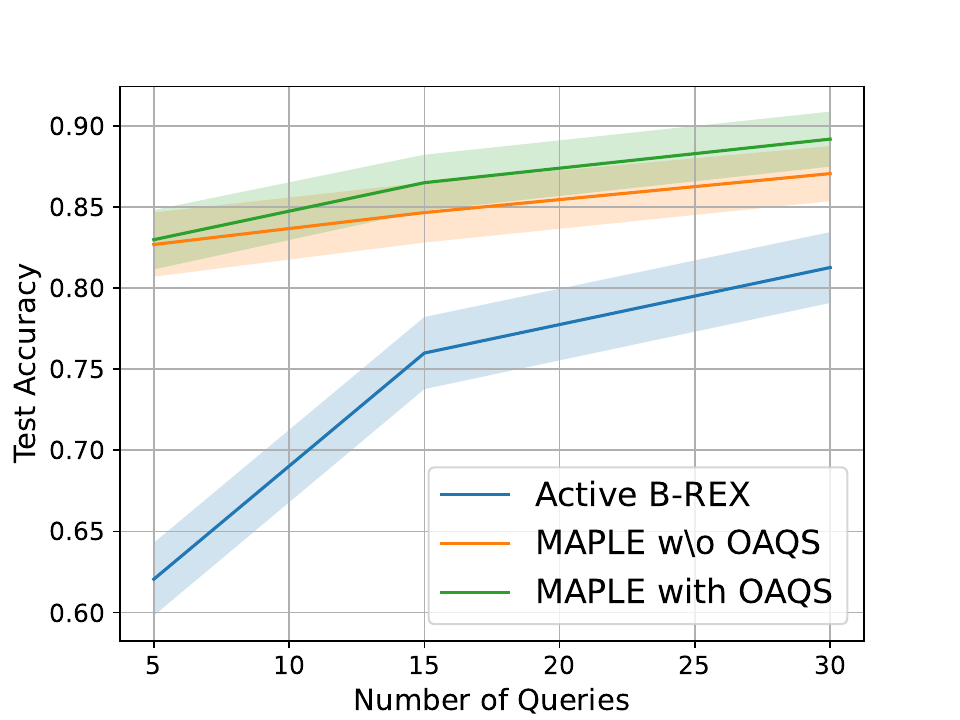}
        \caption*{(a) Test accuracy (OSM Routing)}
    \end{minipage}
    \begin{minipage}[b]{0.30\textwidth}
        \centering
        \includegraphics[width=0.8\textwidth]{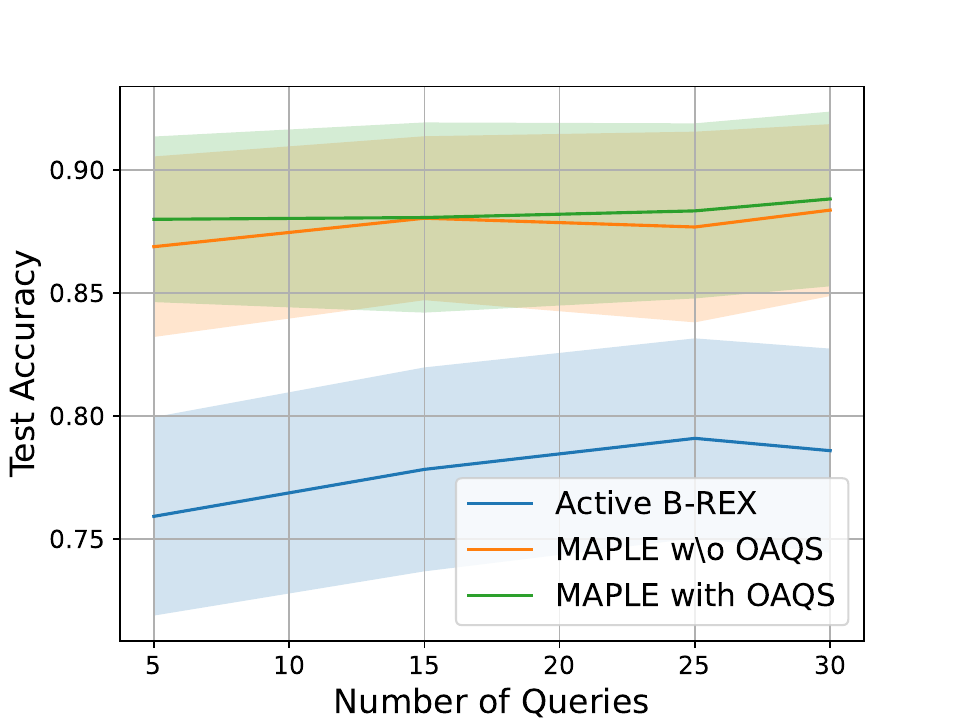}
        \caption*{(b) Test accuracy (HomeGrid)}
    \end{minipage}
    \hfill
        \begin{minipage}[b]{0.36\textwidth}
        \centering
        \tiny
        \begin{tabular}{@{}lcc|cc@{}}
        
            \toprule
            & \multicolumn{2}{c|}{\textbf{OSM Routing}} & \multicolumn{2}{c}{\textbf{HomeGrid}} \\ 
            \textbf{Model} & \textbf{$Y_0$} & \textbf{$Y_1$} & \textbf{$Y_0$} & \textbf{$Y_1$} \\ \midrule
            Mistral-7B-Ins. & 0.30 & 0.65 & 0.17 & 0.77 \\
            GPT-4o  & \textbf{0.84} & \textbf{0.43} & 0.53 & 0.58 \\
            GPT-4o-mini & 0.56 & 0.48 & 0.36 & 0.75 \\
            Gemini-1.5-Pro & 0.62 & 0.72 & \textbf{0.50} & \textbf{0.87} \\
            Mistral-Large-2 & 0.50 & 0.78 & 0.49 & 0.84 \\ \bottomrule
        \end{tabular}
        \caption*{(c) Performance of different models for in-context query selection (See Propositions 3 and 6)}
    \end{minipage}
    \vspace{-5pt}
    \caption{Efficacy of Oracle-guided Active Query Selection (OAQS).}
    \label{fig:combined}
    \vspace{-8pt}
\end{figure*}

We use three key metrics for evaluation: 1) the cosine distance between inferred preference weights (MAP of the distribution) and ground truth preference weights; 2) preference prediction accuracy, which evaluates the model's ability to generalize and accurately predict human preferences from an unseen set of trajectories; and 3) the policy cost difference, which compares the true cost of policies calculated using the ground truth preference function and the learned preference function.

\paragraph{Impact of linguistic feedback}
Figure\,4a-c presents the results of the OSM routing domain experiments. In this experiment, we did not apply OAQS; instead, we selected queries randomly from the dataset to isolate the impact of language. Several noteworthy insights emerge from the results. First, we observe that MAPLE outperforms B-REX on both the natural and clear datasets, demonstrating the effectiveness of integrating complex language feedback with conventional feedback. Additionally, as feedback increases, B-REX's accuracy begins to approach that of MAPLE. This suggests that MAPLE is particularly advantageous when feedback is limited, such as in online settings where the agent must quickly infer rewards.

Examining the cosine distance offers further insight. Language alone appears almost sufficient to align the reward angle, as the cosine distance remains static despite the increasing number of queries. This suggests that preference feedback is more effective for calibrating the magnitude of the preference vector rather than its direction. In contrast, while B-REX achieves good accuracy with large amounts of feedback, it seems to exhibit significant misalignment, which could suggest overfitting and potential failure in out-of-distribution scenarios. Lastly, we evaluated existing publicly available models and found that both GPT-4o and GPT-4o-mini outperformed other models. However, the small local model (Mistral-7B Instruct) proved to be competitive, so we used it to generate all the results shown in Figures 4a, 4b, 4d, and 4e.

Figure\,4d-f shows the results of the HomeGrid experiments. In this environment, we observe that natural instructions do affect performance, but MAPLE still significantly outperforms B-REX in both datasets. Notably, the Mistral-Large-2 models surpassed B-REX by a wide margin, achieving nearly one-third of the cost difference. Surprisingly, GPT-4-mini performed poorly, with a worse cost difference than B-REX. This is due to its inference of highly misaligned preference weights for certain instructions. In this environment, we also see that most of the angle alignment was done using the language feedback and B-REX remains highly misaligned even after 30 feedback. 

\paragraph{Impact of OAQS}

The results of the Oracle-Guided Active Query Selection (OAQS) using an LLM as an oracle are shown in Figure 5. In the routing environment, the Active Query Success Rate (AQSR) is approximately 0.64, while in the HomeGrid environment, it is 0.46. We first evaluated the capability of various models for in-context query selection (Figure\,5c) using a dataset of 500 queries. The Mistral-7B model, used in the previous experiment, failed to meet the condition $Y_0+Y_1>1$ in both environments. The Gemini-1.5-Pro model showed the best overall performance among publicly available models and was used to generate Figures\,5a and 5b.

Figures\,5a and 5b compare the test accuracy of Active B-REX with MAPLE, both with and without OAQS. In both environments, MAPLE with OAQS achieved the highest performance, with a significant margin in the OSM routing environment. We also calculated the Query Success Rate (QSR) for all three algorithms: 0.43 for Active B-REX, 0.43 for MAPLE without OAQS, and 0.58 for MAPLE with OAQS in the routing domain. The QSR was lower than the AQSR for the top-query selection strategy due to a violation of the independence assumption, suggesting that the variance ratio is more likely to select more challenging queries. We refer to this experimental metric as the Effective Query Success Rate (EQSR). Based on Proposition 3, the QSR for MAPLE with OAQS should be 0.77, but it was observed to be lower for the same reason. Replacing AQSR with EQSR in Proposition 3 gives us a value of 0.59, which closely matches the experimental value. Therefore, we conclude that EQSR is a more practical metric for estimating a model's success based on $Y_0$ and $Y_1$. This phenomenon is also observed in HomeGrid. Finally, in the HomeGrid environment, the overall EQSR was low (around 0.2); therefore, even with OAQS, we saw an increase of 2-3 feedback signals after 30 queries, which was not enough to create a large margin and therefore we see only a modest difference between MAPLE with and without OAQS. 
\vspace{-7pt}
\section{Conclusions and Future Works} 

We introduced MAPLE, a framework for active preference learning guided by large language models (LLMs). Our experiments in the OpenStreetMap Routing and HomeGrid environments demonstrated that incorporating language descriptions and explanations significantly improves preference alignment, and that LLM-guided active query selection enhances sample efficiency while reducing the burden on users. Future work could extend MAPLE to more complex environments and tasks, explore different types of linguistic feedback, and conduct user studies to evaluate its usability and effectiveness in real-world applications.

\section{Acknowledgments}
This research was supported in part by the U.S.~Army DEVCOM 
Analysis~Center~(DAC) under contract number W911QX23D0009, and by the National Science Foundation under grants 2321786, 2326054, and 2416459.

\bibliography{aaai25}

\begin{thebibliography}{51}
\providecommand{\natexlab}[1]{#1}

\bibitem[{Abbeel and Ng(2004)}]{abbeel2004apprenticeship}
Abbeel, P.; and Ng, A.~Y. 2004.
\newblock Apprenticeship learning via inverse reinforcement learning.
\newblock In \emph{Proceedings of the 21st International Conference on Machine learning}.

\bibitem[{Achiam et~al.(2023)Achiam, Adler, Agarwal, Ahmad, Akkaya, Aleman, Almeida, Altenschmidt, Altman, Anadkat et~al.}]{achiam2023gpt}
Achiam, J.; Adler, S.; Agarwal, S.; Ahmad, L.; Akkaya, I.; Aleman, F.~L.; Almeida, D.; Altenschmidt, J.; Altman, S.; Anadkat, S.; et~al. 2023.
\newblock GPT-4 technical report.
\newblock \emph{arXiv preprint arXiv:2303.08774}.

\bibitem[{Basu, Singhal, and Dragan(2018)}]{Basu2018LearningFR}
Basu, C.; Singhal, M.; and Dragan, A.~D. 2018.
\newblock Learning from richer human guidance: Augmenting comparison-based learning with feature queries.
\newblock In \emph{13th International Conference on Human-Robot Interaction}, 132--140.

\bibitem[{Biyik(2022)}]{biyik2022learning}
Biyik, E. 2022.
\newblock \emph{Learning preferences for interactive autonomy}.
\newblock Ph.D. thesis, Stanford University.

\bibitem[{Biyik et~al.(2019)Biyik, Palan, Landolfi, Losey, and Sadigh}]{biyik2019asking}
Biyik, E.; Palan, M.; Landolfi, N.~C.; Losey, D.~P.; and Sadigh, D. 2019.
\newblock Asking easy questions: A user-friendly approach to active reward learning.
\newblock In \emph{Proceedings of the 3rd Annual Conference on Robot Learning}, 1177--1190.

\bibitem[{Bobu et~al.(2021)Bobu, Paxton, Yang, Sundaralingam, Chao, Cakmak, and Fox}]{bobu2021learning}
Bobu, A.; Paxton, C.; Yang, W.; Sundaralingam, B.; Chao, Y.-W.; Cakmak, M.; and Fox, D. 2021.
\newblock Learning perceptual concepts by bootstrapping from human queries.
\newblock \emph{arXiv preprint arXiv:2111.05251}.

\bibitem[{Bradley and Terry(1952)}]{Bradley1952RankAO}
Bradley, R.~A.; and Terry, M.~E. 1952.
\newblock Rank Analysis of Incomplete Block Designs: I. The Method of Paired Comparisons.
\newblock \emph{Biometrika}, 39: 324.

\bibitem[{Brown, Goo, and Niekum(2019)}]{brown2019drex}
Brown, D.~S.; Goo, W.; and Niekum, S. 2019.
\newblock Better-than-demonstrator imitation learning via automatically-ranked demonstrations.
\newblock In \emph{3rd Annual Conference on Robot Learning}, 330--359.

\bibitem[{Brown et~al.(2019)Brown, Goo, Prabhat, and Niekum}]{browngoo2019trex}
Brown, D.~S.; Goo, W.; Prabhat, N.; and Niekum, S. 2019.
\newblock Extrapolating beyond suboptimal demonstrations via inverse reinforcement learning from observations.
\newblock In \emph{36th International Conference on Machine Learning}, 783--792.

\bibitem[{Brown et~al.(2020)Brown, Niekum, Coleman, and Srinivasan}]{brown2020safe}
Brown, D.~S.; Niekum, S.; Coleman, R.; and Srinivasan, R. 2020.
\newblock Safe imitation learning via fast Bayesian reward inference from preferences.
\newblock In \emph{37th International Conference on Machine Learning}, 1165--1177.

\bibitem[{Brown, Schneider, and Niekum(2021)}]{Brown2021ValueAV}
Brown, D.~S.; Schneider, J.~J.; and Niekum, S. 2021.
\newblock Value alignment verification.
\newblock In \emph{38th International Conference on Machine Learning}, 1105--1115.

\bibitem[{Bucker et~al.(2023)Bucker, Figueredo, Haddadin, Kapoor, Ma, Vemprala, and Bonatti}]{bucker2023latte}
Bucker, A.; Figueredo, L. F.~C.; Haddadin, S.; Kapoor, A.; Ma, S.; Vemprala, S.; and Bonatti, R. 2023.
\newblock {LATTE:} LAnguage Trajectory TransformEr.
\newblock In \emph{{IEEE} International Conference on Robotics and Automation}, 7287--7294.

\bibitem[{Chebotar et~al.(2021)Chebotar, Hausman, Lu, Xiao, Kalashnikov, Varley, Irpan, Eysenbach, Julian, Finn et~al.}]{chebotar2021actionable}
Chebotar, Y.; Hausman, K.; Lu, Y.; Xiao, T.; Kalashnikov, D.; Varley, J.; Irpan, A.; Eysenbach, B.; Julian, R.; Finn, C.; et~al. 2021.
\newblock Actionable models: Unsupervised offline reinforcement learning of robotic skills.
\newblock \emph{arXiv preprint arXiv:2104.07749}.

\bibitem[{Chevalier{-}Boisvert et~al.(2023)Chevalier{-}Boisvert, Dai, Towers, Perez{-}Vicente, Willems, Lahlou, Pal, Castro, and Terry}]{MinigridMiniworld23}
Chevalier{-}Boisvert, M.; Dai, B.; Towers, M.; Perez{-}Vicente, R.; Willems, L.; Lahlou, S.; Pal, S.; Castro, P.~S.; and Terry, J. 2023.
\newblock Minigrid {\&} Miniworld: Modular {\&} customizable reinforcement learning environments for goal-oriented tasks.
\newblock In \emph{Advances in Neural Information Processing Systems 36}.

\bibitem[{Cui et~al.(2023)Cui, Karamcheti, Palleti, Shivakumar, Liang, and Sadigh}]{cui2023no}
Cui, Y.; Karamcheti, S.; Palleti, R.; Shivakumar, N.; Liang, P.; and Sadigh, D. 2023.
\newblock ``No, to the Right'' -- Online language corrections for robotic manipulation via shared autonomy.
\newblock \emph{arXiv preprint arXiv:2301.02555}.

\bibitem[{Dietterich(2017)}]{dietterich2017steps}
Dietterich, T.~G. 2017.
\newblock Steps toward robust artificial intelligence.
\newblock \emph{{AI} Magazine}, 38(3): 3--24.

\bibitem[{Gal and Ghahramani(2016)}]{gal2016dropout}
Gal, Y.; and Ghahramani, Z. 2016.
\newblock Dropout as a Bayesian approximation: Representing model uncertainty in deep learning.
\newblock In \emph{33nd International Conference on Machine Learning}, 1050--1059.

\bibitem[{Guan, Sreedharan, and Kambhampati(2022)}]{guan2022leveraging}
Guan, L.; Sreedharan, S.; and Kambhampati, S. 2022.
\newblock Leveraging approximate symbolic models for reinforcement learning via skill diversity.
\newblock \emph{arXiv preprint arXiv:2202.02886}.

\bibitem[{Guan, Valmeekam, and Kambhampati(2022)}]{guan2022relative}
Guan, L.; Valmeekam, K.; and Kambhampati, S. 2022.
\newblock Relative behavioral attributes: Filling the gap between symbolic goal specification and reward learning from human preferences.
\newblock \emph{arXiv preprint arXiv:2210.15906}.

\bibitem[{Guan et~al.(2021)Guan, Verma, Guo, Zhang, and Kambhampati}]{guan2021widening}
Guan, L.; Verma, M.; Guo, S.~S.; Zhang, R.; and Kambhampati, S. 2021.
\newblock Widening the pipeline in human-guided reinforcement learning with explanation and context-aware data augmentation.
\newblock \emph{Advances in Neural Information Processing Systems}, 34: 21885--21897.

\bibitem[{Guo et~al.(2022)Guo, Zou, Zuo, Wang, Ji, Li, and Cheng}]{guo2022generating}
Guo, C.; Zou, S.; Zuo, X.; Wang, S.; Ji, W.; Li, X.; and Cheng, L. 2022.
\newblock Generating diverse and natural 3D human motions from text.
\newblock In \emph{IEEE/CVF Conference on Computer Vision and Pattern Recognition}, 5152--5161.

\bibitem[{Icarte et~al.(2022)Icarte, Klassen, Valenzano, and McIlraith}]{icarte2022reward}
Icarte, R.~T.; Klassen, T.~Q.; Valenzano, R.; and McIlraith, S.~A. 2022.
\newblock Reward machines: Exploiting reward function structure in reinforcement learning.
\newblock \emph{Journal of Artificial Intelligence Research}, 73: 173--208.

\bibitem[{Illanes et~al.(2020)Illanes, Yan, Icarte, and McIlraith}]{illanes2020symbolic}
Illanes, L.; Yan, X.; Icarte, R.~T.; and McIlraith, S.~A. 2020.
\newblock Symbolic plans as high-level instructions for reinforcement learning.
\newblock In \emph{30th International Conference on Automated Planning and Scheduling}, 540--550.

\bibitem[{Lee and Popovi{\'c}(2010)}]{lee2010learning}
Lee, S.~J.; and Popovi{\'c}, Z. 2010.
\newblock Learning behavior styles with inverse reinforcement learning.
\newblock \emph{ACM transactions on graphics}, 29(4): 1--7.

\bibitem[{Lin et~al.(2023)Lin, Du, Watkins, Hafner, Abbeel, Klein, and Dragan}]{lin2023learning}
Lin, J.; Du, Y.; Watkins, O.; Hafner, D.; Abbeel, P.; Klein, D.; and Dragan, A. 2023.
\newblock Learning to model the world with language.
\newblock \emph{arXiv preprint arXiv:2308.01399}.

\bibitem[{Lin et~al.(2022)Lin, Fried, Klein, and Dragan}]{lin2022inferring}
Lin, J.; Fried, D.; Klein, D.; and Dragan, A. 2022.
\newblock Inferring rewards from language in context.
\newblock \emph{arXiv preprint arXiv:2204.02515}.

\bibitem[{Lou et~al.(2024)Lou, Zhang, Wang, Huang, and Du}]{lou2024safe}
Lou, X.; Zhang, J.; Wang, Z.; Huang, K.; and Du, Y. 2024.
\newblock Safe reinforcement learning with free-form natural language constraints and pre-trained language models.
\newblock \emph{arXiv preprint arXiv:2401.07553}.

\bibitem[{Luo et~al.(2020)Luo, Soeseno, Chen, and Chen}]{luo2020carl}
Luo, Y.-S.; Soeseno, J.~H.; Chen, T. P.-C.; and Chen, W.-C. 2020.
\newblock CARL: Controllable agent with reinforcement learning for quadruped locomotion.
\newblock \emph{ACM Transactions on Graphics}, 39(4): 38--1.

\bibitem[{Lyu et~al.(2019)Lyu, Yang, Liu, and Gustafson}]{lyu2019sdrl}
Lyu, D.; Yang, F.; Liu, B.; and Gustafson, S. 2019.
\newblock SDRL: interpretable and data-efficient deep reinforcement learning leveraging symbolic planning.
\newblock In \emph{33rd AAAI Conference on Artificial Intelligence}, 2970--2977.

\bibitem[{Ma et~al.(2023)Ma, Liang, Wang, Huang, Bastani, Jayaraman, Zhu, Fan, and Anandkumar}]{ma2023eureka}
Ma, Y.~J.; Liang, W.; Wang, G.; Huang, D.-A.; Bastani, O.; Jayaraman, D.; Zhu, Y.; Fan, L.; and Anandkumar, A. 2023.
\newblock Eureka: Human-level reward design via coding large language models.
\newblock \emph{arXiv preprint arXiv:2310.12931}.

\bibitem[{Mahmud, Saisubramanian, and Zilberstein(2023)}]{rev}
Mahmud, S.; Saisubramanian, S.; and Zilberstein, S. 2023.
\newblock Explanation-guided reward alignment.
\newblock In \emph{32nd International Joint Conference on Artificial Intelligence}, 473--482.

\bibitem[{Ng and Russell(2000)}]{ng2000algorithms}
Ng, A.~Y.; and Russell, S.~J. 2000.
\newblock Algorithms for inverse reinforcement learning.
\newblock In \emph{17th International Conference on Machine Learning}, 663--670.

\bibitem[{{OpenStreetMap Contributors}(2017)}]{OpenStreetMap}
{OpenStreetMap Contributors}. 2017.
\newblock {Planet dump retrieved from https://planet.osm.org }.
\newblock \url{ https://www.openstreetmap.org }.

\bibitem[{Peng et~al.(2018)Peng, Kanazawa, Malik, Abbeel, and Levine}]{peng2018sfv}
Peng, X.~B.; Kanazawa, A.; Malik, J.; Abbeel, P.; and Levine, S. 2018.
\newblock SFV: Reinforcement learning of physical skills from videos.
\newblock \emph{ACM Transactions On Graphics}, 37(6): 178:1--178:14.

\bibitem[{Peng et~al.(2021)Peng, Ma, Abbeel, Levine, and Kanazawa}]{2021-TOG-AMP}
Peng, X.~B.; Ma, Z.; Abbeel, P.; Levine, S.; and Kanazawa, A. 2021.
\newblock AMP: Adversarial motion priors for stylized physics-based character control.
\newblock \emph{ACM Transactions On Graphics}, 40(4): 144:1--144:20.

\bibitem[{Rafailov et~al.(2024)Rafailov, Sharma, Mitchell, Manning, Ermon, and Finn}]{rafailov2024direct}
Rafailov, R.; Sharma, A.; Mitchell, E.; Manning, C.~D.; Ermon, S.; and Finn, C. 2024.
\newblock Direct preference optimization: Your language model is secretly a reward model.
\newblock \emph{Advances in Neural Information Processing Systems}, 36.

\bibitem[{Ramachandran and Amir(2007)}]{ramachandran2007bayesian}
Ramachandran, D.; and Amir, E. 2007.
\newblock Bayesian inverse reinforcement learning.
\newblock In \emph{20th International Joint Conference on Artifical intelligence}, 2586--2591.

\bibitem[{Sadigh et~al.(2017)Sadigh, Dragan, Sastry, and Seshia}]{sadigh2017active}
Sadigh, D.; Dragan, A.~D.; Sastry, S.~S.; and Seshia, S.~A. 2017.
\newblock Active preference-based learning of reward functions.
\newblock In \emph{Robotics: Science and Systems XIII}.

\bibitem[{Silver et~al.(2022)Silver, Athalye, Tenenbaum, Lozano-Perez, and Kaelbling}]{silver2022learning}
Silver, T.; Athalye, A.; Tenenbaum, J.~B.; Lozano-Perez, T.; and Kaelbling, L.~P. 2022.
\newblock Learning neuro-symbolic skills for bilevel planning.
\newblock \emph{arXiv preprint arXiv:2206.10680}.

\bibitem[{Soni et~al.(2022)Soni, Thakur, Sreedharan, Guan, Verma, Marquez, and Kambhampati}]{soni2022towards}
Soni, U.; Thakur, N.; Sreedharan, S.; Guan, L.; Verma, M.; Marquez, M.; and Kambhampati, S. 2022.
\newblock Towards customizable reinforcement learning agents: Enabling preference specification through online vocabulary expansion.
\newblock \emph{arXiv preprint arXiv:2210.15096}.

\bibitem[{Sontakke et~al.(2024)Sontakke, Zhang, Arnold, Pertsch, Biyik, Sadigh, Finn, and Itti}]{sontakke2024roboclip}
Sontakke, S.; Zhang, J.; Arnold, S.; Pertsch, K.; Biyik, E.; Sadigh, D.; Finn, C.; and Itti, L. 2024.
\newblock RoboCLIP: One demonstration is enough to learn robot policies.
\newblock \emph{Advances in Neural Information Processing Systems}, 36.

\bibitem[{Tevet et~al.(2022)Tevet, Raab, Gordon, Shafir, Cohen-Or, and Bermano}]{tevet2022human}
Tevet, G.; Raab, S.; Gordon, B.; Shafir, Y.; Cohen-Or, D.; and Bermano, A.~H. 2022.
\newblock Human motion diffusion model.
\newblock \emph{arXiv preprint arXiv:2209.14916}.

\bibitem[{Tian et~al.(2024)Tian, Li, Li, Ran, Ning, and Tiwari}]{tian2024survey}
Tian, S.; Li, L.; Li, W.; Ran, H.; Ning, X.; and Tiwari, P. 2024.
\newblock A survey on few-shot class-incremental learning.
\newblock \emph{Neural Networks}, 169: 307--324.

\bibitem[{Tien et~al.(2024)Tien, Yang, Jun, Russell, Dragan, and Biyik}]{tien2024optimizing}
Tien, J.; Yang, Z.; Jun, M.; Russell, S.~J.; Dragan, A.; and Biyik, E. 2024.
\newblock Optimizing robot behavior via comparative language feedback.
\newblock In \emph{3rd HRI Workshop on Human-Interactive Robot Learning}.

\bibitem[{Wang et~al.(2024)Wang, Sun, Zhang, Xian, Biyik, Held, and Erickson}]{wang2024rl}
Wang, Y.; Sun, Z.; Zhang, J.; Xian, Z.; Biyik, E.; Held, D.; and Erickson, Z. 2024.
\newblock {RL-VLM-F}: Reinforcement learning from vision language foundation model feedback.
\newblock \emph{arXiv preprint arXiv:2402.03681}.

\bibitem[{Wang et~al.(2017)Wang, Merel, Reed, de~Freitas, Wayne, and Heess}]{wang2017robust}
Wang, Z.; Merel, J.~S.; Reed, S.~E.; de~Freitas, N.; Wayne, G.; and Heess, N. 2017.
\newblock Robust imitation of diverse behaviors.
\newblock \emph{Advances in Neural Information Processing Systems}, 30.

\bibitem[{Yu et~al.(2023)Yu, Gileadi, Fu, Kirmani, Lee, Arenas, Chiang, Erez, Hasenclever, Humplik et~al.}]{yu2023language}
Yu, W.; Gileadi, N.; Fu, C.; Kirmani, S.; Lee, K.-H.; Arenas, M.~G.; Chiang, H.-T.~L.; Erez, T.; Hasenclever, L.; Humplik, J.; et~al. 2023.
\newblock Language to rewards for robotic skill synthesis.
\newblock \emph{arXiv preprint arXiv:2306.08647}.

\bibitem[{Zhang et~al.(2022)Zhang, Bansal, Hao, Hiranaka, Gao, Wang, Mart{\'\i}n-Mart{\'\i}n, Fei-Fei, and Wu}]{zhangdual}
Zhang, R.; Bansal, D.; Hao, Y.; Hiranaka, A.; Gao, J.; Wang, C.; Mart{\'\i}n-Mart{\'\i}n, R.; Fei-Fei, L.; and Wu, J. 2022.
\newblock A dual representation framework for robot learning with human guidance.
\newblock In \emph{6th Annual Conference on Robot Learning}, 738--750.

\bibitem[{Zhou and Dragan(2018)}]{zhou2018cost}
Zhou, A.; and Dragan, A.~D. 2018.
\newblock Cost functions for robot motion style.
\newblock In \emph{2018 IEEE/RSJ International Conference on Intelligent Robots and Systems}, 3632--3639. IEEE.

\bibitem[{Ziebart et~al.(2008)Ziebart, Maas, Bagnell, and Dey}]{ziebart2008maximum}
Ziebart, B.~D.; Maas, A.~L.; Bagnell, J.~A.; and Dey, A.~K. 2008.
\newblock Maximum entropy inverse reinforcement learning.
\newblock In \emph{Proceedings of the 23rd AAAI Conference on Artificial Intelligence}, 1433--1438.

\bibitem[{Zilberstein(2015)}]{zilberstein2015building}
Zilberstein, S. 2015.
\newblock Building strong semi-autonomous systems.
\newblock In \emph{29th AAAI Conference on Artificial Intelligence}, 4088--4092.

\end{thebibliography}
\end{document}